\documentclass[letterpaper]{article} 
\usepackage{aaai24}  
\usepackage{times}  
\usepackage{helvet}  
\usepackage{courier}  
\usepackage[hyphens]{url}  
\usepackage{graphicx} 
\usepackage{natbib}  
\usepackage{caption} 
\usepackage{amsmath} 
\usepackage{array}
\usepackage{footnote}
\usepackage{makecell, multirow}
\usepackage{booktabs,makecell, multirow, tabularx, enumitem}
\makesavenoteenv{tabular}
\usepackage{multirow}
\usepackage{algorithm}
\usepackage{algorithmic}
\usepackage[title]{appendix}

\usepackage{newfloat}
\usepackage{listings}
\DeclareCaptionStyle{ruled}{labelfont=normalfont,labelsep=colon,strut=off} 
\floatstyle{ruled}
\newfloat{listing}{tb}{lst}{}
\floatname{listing}{Listing}
\newcommand{\ourmethod}[1]{}
\renewcommand{\ourmethod}[1]{\texttt{GADY}}
\title{GADY: Unsupervised Anomaly Detection on Dynamic Graphs}
\author {
    Shiqi Lou\textsuperscript{\rm 1},
    Qingyue Zhang\textsuperscript{\rm 2},
    Shujie Yang\textsuperscript{\rm 2}
    Yuyang Tian\textsuperscript{\rm 1}
    Zhaoxuan Tan\textsuperscript{\rm 3}
    Minnan Luo\textsuperscript{\rm 1}
}
\affiliations {
    \textsuperscript{\rm 1}Xi'an Jiaotong University, 
    \textsuperscript{\rm 2}Tsinghua University,
    \textsuperscript{\rm 3}University of Notre Dame\\
    lousqi@stu.xjtu.edu.cn,\{zhangqy23, yangsj23\}@mails.tsinghua.edu.cn, 
    2191414578@stu.xjtu.edu.cn, \\
    ztan3@nd.edu, minnluo@xjtu.edu.cn
}

\begin{document}

\maketitle

\begin{abstract}
Anomaly detection on dynamic graphs refers to detecting entities whose behaviors obviously deviate from the norms observed within graphs and their temporal information. This field has drawn increasing attention due to its application in finance, network security, social networks, and more. However, existing methods face two challenges: dynamic structure constructing challenge -  difficulties in capturing graph structure with complex time information and negative sampling challenge - unable to construct excellent negative samples for unsupervised learning. To address these challenges, we propose Unsupervised \textbf{G}enerative \textbf{A}nomaly Detection on \textbf{Dy}namic Graphs (\textbf{GADY}). To tackle the first challenge, we propose a \textit{continuous dynamic graph} model to capture the fine-grained information, which breaks the limit of existing discrete methods. Specifically, we employ a message-passing framework combined with positional features to get edge embeddings, which are decoded to identify anomalies. For the second challenge, we pioneer the use of Generative Adversarial Networks to generate negative interactions. Moreover, we design a loss function to alter the training goal of the generator while ensuring the diversity and quality of generated samples. Extensive experiments demonstrate that our proposed \ourmethod{} significantly outperforms the previous state-of-the-art method on three real-world datasets. Supplementary experiments further validate the effectiveness of our model design and the necessity of each module.
\end{abstract}

\section{Introduction}





Anomaly detection on dynamic graphs, has become an important task given its numerous applications, such as social media spammer detection \cite{Ye2015DiscoveringOS}, fraudulent transaction detection \cite{financial}, and network intrusion detection \cite{network}. Capturing the anomaly on dynamic graphs can help us better understand and capture the evolution of social networks \cite{Ye2015DiscoveringOS}, financial transactions \cite{financial}, and disease diagnosis \cite{disease} with fine-grained time information. For example, in financial transactions, if two entities conduct a transaction and they are in two transaction collectives that had no transaction records before, this transaction is very suspicious, which highlights the importance of temporal information in detecting anomalies.



%

\begin{figure}[t]
    \centering
    \includegraphics[width=1\linewidth]{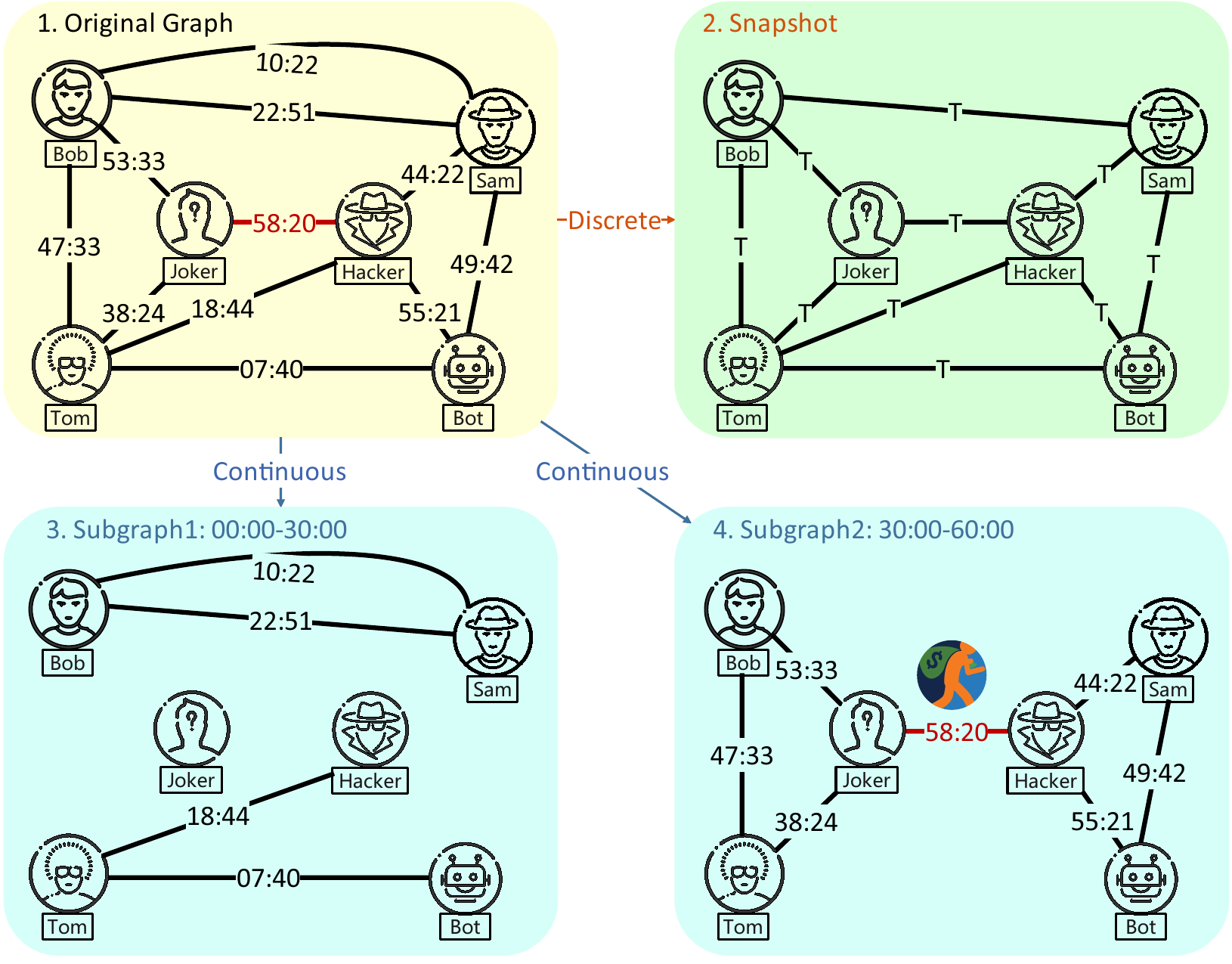}
    \caption{Example of the superiority of anomaly detection in a continuous dynamic graph. 
    When detecting the anomaly edge of 58:20 in the dynamic original graph, discrete dynamic graph methods fall short, while continuous dynamic methods can easily distinguish this anomaly behavior.}
    \label{teaser}
\end{figure}

Recently, many studies have been done in this area \cite{addgraph,strgnn,taddy}. They usually divide the entire temporal graph into several snapshots and conduct anomaly detection based on this graph format. Addgraph \cite{addgraph} uses GCN to extract information about graph structure from the slices and then using the GRU-attention module to construct long- and short-term relationships between the slices.  StrGNN \cite{strgnn} extracted h-hop closed subgraphs centered on edges, and then used GCN and GRU to model the structural information on snapshots and the correlation between snapshots respectively. Taddy \cite{taddy} uses a transformer to process diffusion-based spatial encoding, distance-based spatial encoding and relative time encoding, and then obtain edge representation through pooling layer, thus obtaining anomaly score.

Although much research progress has been achieved in dynamic graph anomaly detection, they are weak in two aspects: dynamic structure constructing challenge -  difficulties in capturing graph structure with complex time information and negative sampling challenge - unable to construct excellent negative samples for unsupervised learning. For the dynamic structure constructing challenge, anomaly detection usually needs entity embedding through the encoder part, which are then decoded as labels. Some research works propose to use discrete dynamic methods for modeling \cite{strgnn,taddy}. Specifically, they cut dynamic graphs into several slices, which can be captured by Graph Neural Networks (GNNs) or similar methods. Then they uses Gated Recurrent Unit (GRU) and similar methods to construct the evolutionary process of slices \cite{addgraph,strgnn,taddy}. These methods are able to plug in well-developed GNN methods to boost the model performance. However, using this method inevitably results in two kinds of information loss. Firstly, it ignores repeated edges in the slice time interval, which will cause GNN to fail to capture complete interaction information, making the generated entity embedding defective. Secondly, it ignores the time difference within the same slice, making it difficult for the model to capture time dependencies. Both kinds of information loss would hinder the anomaly detection performance on dynamic graphs.

In addition to the above problem, research on anomaly detection on dynamic graphs also faces the negative sampling challenge: unable to construct excellent negative samples for unsupervised learning, which is, on the existing dynamic graph datasets, there are no real anomalous samples to help us conduct supervised training. To solve this problem, some people propose a context-dependent negative sampling method to artificially construct negative samples by rules \cite{strgnn} for unsupervised learning. Specifically, for each edge, they randomly keep the head node or tail node and replace the unreserved nodes with other irrelevant nodes, thus obtaining negative samples for training. However, the quality of negative samples in the training phase has a crucial impact on the performance of the model. Ideally, the generated negative samples will enable the model to efficiently learn diverse anomaly patterns during the training phase. This puts forward two requirements for negative samples: 1. Abnormal samples should be of high quality; 2. Abnormal samples need to be diverse. The proposed method has limited performance in these two criteria.



To break the above two challenges, we propose a general model for continuous dynamic graph anomaly detection. Our main idea is to use an anomaly generator to obtain diverse and high-quality negative samples to help training, and to use a continuous dynamic graph model to perform anomaly detection. Through adversarial training of the two modules, an excellent discriminator can finally be obtained, thus achieving better performance on anomaly detection. Specifically, to break the limitation of the existing discrete dynamic graph detection methods, we use message-passing continuous dynamic models combined with positional features \cite{pint} to capture complex dynamic graph. Also, we use anomaly generator to generate diverse negative edges with high quality. To achieve this, we propose a novel loss function designed for dynamic graphs anomaly detection to help generator find its specific training goal. 
We follow the evaluation settings proposed by \citet{taddy} for fair comparison. Experimental results show that our proposed \ourmethod{} outperforms the current state of the art significantly on three benchmarks, demonstrating the superiority of our design choice. We also investigate the potential of current continuous dynamic graphs for anomaly detection. Detailed supplementary experiments about GAN module and generator design also validate the necessity of anomaly generator and the other components in \ourmethod{}. 
In summary, our main contributions are as follows:
\begin{itemize}
\item 
We identify the drawbacks of existing discrete dynamic graph anomaly detection methods and further explore the potential of continuous dynamic graph models to address existing dynamic structure constructing challenge.
\item We pioneer the use of the GAN network for anomaly detection on dynamic graphs and demonstrate the effectiveness of the GAN model in improving anomaly detection capabilities.

\item Extensive experiments on three datasets show that our model has achieved a maximum improvement of 14.6\% in anomaly detection compared with existing methods, and achieved extraordinary performance in all anomaly proportions of all datasets. Extensive supplementary experiments again demonstrate the necessity of anomaly generators and the superiority of model settings.
\end{itemize}
\section{Related Work}
\subsection{Anomaly Detection on Dynamic Graph}




In recent years, anomaly detection has focused on using deep learning methods to model the entire process. \cite{jin2021anemone, zhao2021action,tariq2022towards,zhou2021subtractive,han2021unsupervised,guo2022learning} The processing methods of dynamic graphs can be divided into discrete processing methods and continuous processing methods. Many methods have been proposed using discrete methods in recent years to solve this task, such as  {Addgraph} \cite{addgraph} using GCN to extract graph structure information on slices, and then using GRU-attention module to construct long and short term relationships between slices.  {StrGNN} \cite{strgnn} extracted h-hop closed subgraphs centered on edges, and then used GCN and GRU to model the structural information on snapshots and the correlation between snapshots respectively.  {Taddy} \cite{taddy} uses a transformer to process diffusion-based spatial encoding, distance-based spatial encoding and relative time encoding, and then obtain edge representation through the pooling layer, thus obtaining the anomaly score. For continuous methods, it can be more helpful for anomaly detection with its strong modeling capabilities and sufficient time information. Recently,  {SAD} \cite{SAD} propose to use continuous dynamic methods to detect anomalies using the semi-supervised method, which is different from our work.

\subsection{Generative Adversarial Network in Anomaly Detection}
Nowadays, using GAN to perform anomaly detection is divided into two main branches. The first branch is to use GAN to learn a normal sample distribution and to use the reconstruction error as an anomaly score. For example, AnoGAN \cite{anogan} can capture the latent distribution of normal data through training, and because the behavior patterns of abnormal data and normal data are different, after processing by  {AnoGAN}, the residual between original input and output is usually much larger than normal images, thus detecting anomalies in image.

However, since the original goal of GAN is to generate high-quality data rather than anomaly detection, the anomaly score obtained by using reconstruct loss of GAN for anomaly detection may be suboptimal. Another branch is to use GAN to generate negative samples to assist in model training. For example,  {OCAN} \cite{OCAN} uses LSTM-Autoencoder to learn the representation of normal users, and then modifies the generator so that its training goal becomes generating abnormal data that is complementary to normal data, and obtains a better discriminator after training.  {Fence-GAN} \cite{fencegan} improves the model training effect by defining different loss functions so that the generated samples are located at the edge of normal samples distribution. 

Although our model has similar ideas with  {OCAN}, we use generator to get interactions directly rather than embeddings with different generator structure for its application in unattributed dynamic graphs. In addition, we use a designed loss function to improve our model performance.

\begin{figure*}[htbp]
\begin{center}
\includegraphics[width=1\linewidth]{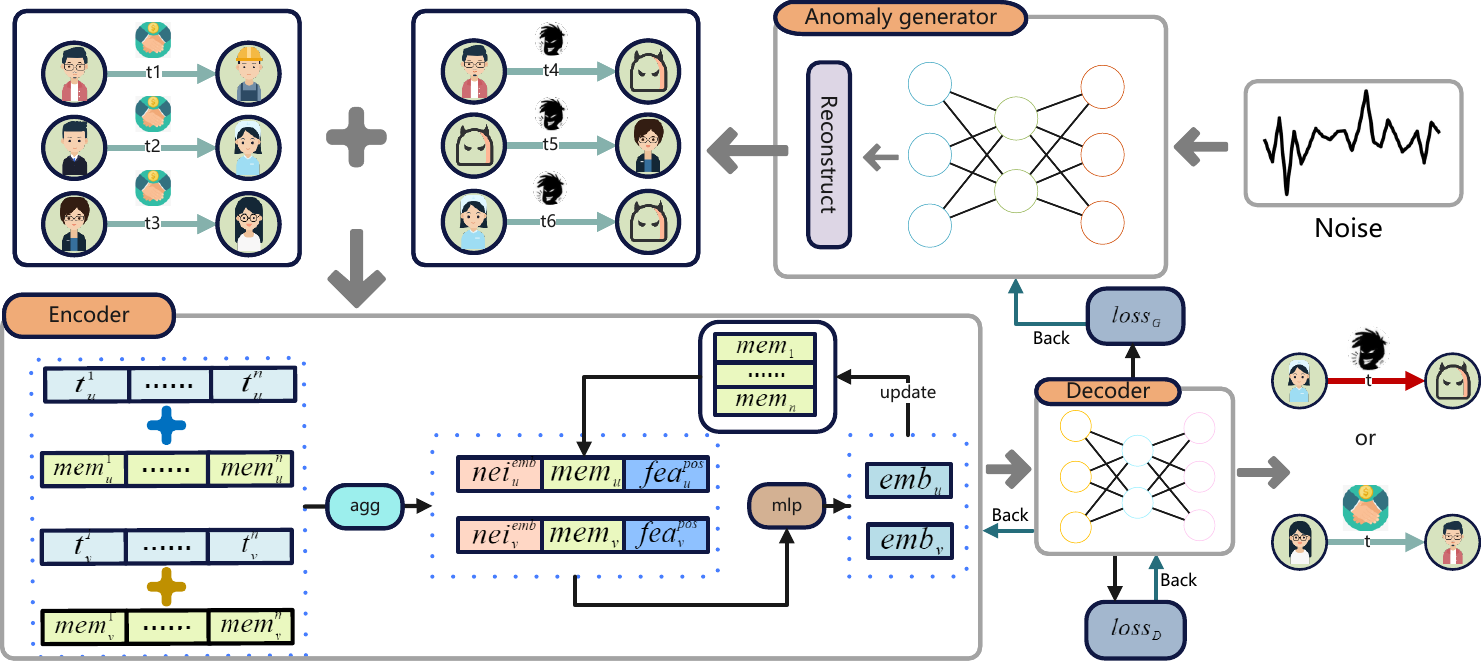}
\end{center}
\caption{The architecture of \ourmethod{} network includes two parts. The first part is the anomaly generator, which generates abnormal interactions through input noise, sorts the generated interactions in time, and inputs them into the encoder part together with the normal interactions. The second part is the discriminator, which contains the encoder and decoder. After getting the interactions, the encoder obtains the edge embedding by continuously aggregating the information of time neighbors (R-agg) for the head node and the tail node. Finally, the embedding of the edge is fed into the decoder. For the decoder, it obtains the anomaly score of the edge through the representation of the head node and the tail node, finally judging whether the target edge is anomalous or normal.}
\label{overview}
\end{figure*}


\section{Methodology}
Given a dynamic graph $G= \{e_{uv}(t)=(u,v,t):u,v\in\mathcal{V})\}$, our goal is to identify a fake interaction $e_{u'v'}(t')=(u',v',t')$ which is different from other edges. To achieve this goal, we design GADY, whose structure is shown in Figure \ref{overview}.
Our model has two parts: anomaly generator G and discriminator D. For the anomaly generator, it accepts the input noise $z_i \in (-1,1)$ and output the fake interactions $G(z_i)$, where $G$ is the generator function. This function is trained through a loss function $L_G(G,D,Z_i)$ where $D$ is the discriminator function.
For the discriminator, it contains two parts: encoder and decoder. For the part of encoder, it is aimed to train a function $E()$, where $E(e_{uv}(t))$ is the output edge embedding. For the part of decoder, it inputs the edge embedding and outputs the anomaly score of the interaction between these two nodes, and then judges whether the edge is an anomalous edge based on the anomaly score. This process can be formulated as $F(emb_u,emb_v)$ where $u,v \in \mathcal{V}$. For the whole discriminator, it can be formulated as $D(e_{uv}(t))=F(E(e_{uv}(t)))$. Also, the loss function of the discriminator is $L_D(G(z_i),x_i,D)$ where $G(z_i)$ and $x_i$ are the generated samples and real samples, respectively.

\subsection{Generator}
\subsubsection{Framework Design} The purpose of the generator is to generate high-quality and diverse negative samples. For the high quality of negative samples, our definition is to be as similar as possible to real samples, but not exactly the same as real samples. Such negative samples can make the model work hard to distinguish true and false samples during the training phase. For the diversity of samples, our definition is to have as many abnormal patterns as possible, so that the model can learn to see more kinds of abnormalities during the training process, thereby improving the detection ability of the model. The whole generator part is designed following these two principles.

The standard input of our generator is random noise such as Gaussian noise or Uniform noise. For the type of input noise, through our experiment, we find that different types of noise are suitable for different datasets. In our model, we use gaussian noise for its superiority. For the generator structure, an immediate question is whether the generator should produce edges or edge embedding. Different from  {OCAN} \cite{OCAN}, Here we have proved through experiments that generating edges is more conducive to the effect of the model \ref{fig:interaction1}, which may be because ignoring the embedding process of encoder for interactions, resulting in bias and sub-optimal outcomes. Moreover, generating fake edges on dynamic graphs still faces some problems. The first is that the edges come in chronological order, and for each batch, the timestamps of the generated edges should be within this range, so as to make the negative samples more similar to the real samples. For the head node and tail node of the edge, we guarantee that the randomly generated nodes are within dataset. Based on the above considerations, we define the model as follows
\begin{equation}
G(z_i) = R(\tanh(f(z_i)))
\end{equation}
where $f$ is defined as a  3-layer multilayer perceptrons and $z_i$ is the input noise. $\tanh()$ is an activation function reconstructing the output to $[0,1]$, and $R$ is a function reconstructing the output embedding of $\tanh()$ to the input form.
Through the processing of multilayer perceptrons, noise is reconstructed into input embedding. Then, we reconstruct the output of the generator to the form of $(u,v,t)$, and then output the anomaly score and judgment through the encoding-decoding process of the discriminator. 

\subsubsection{Loss Function design}
The design of the loss function usually affects the training goal of the model. In order to ensure that the model is trained according to the above two principles, we define the following loss function:

\begin{equation}
\begin{aligned}
L_G(G,D,z_i) &=\frac{1}{N}\sum_{i=1}^{N}\rm{log}(|\alpha -D(G(z_i))|) \\
&+\beta \times \frac{1}{\frac{1}{n}\sum_{k=1}^n(CL^k)}
\end{aligned}
\end{equation}
In this function, $L_G$ is the loss function for the generator, and $G$ and $D$ refer to generator and discriminator respectively. The first part is to ensure the quality of generated samples. $\alpha \in [0,1]$ is expected anomaly score of generated samples from discriminator. Setting $\alpha$ to 0.1 means that the generator hopes that the samples generated by itself will get scores of 0.1 from the discriminator, meaning that although they are judged as normal samples, they still has certain anomalous patterns. By tuning this parameter, we can control the quality of negative samples generated by the generator to adapt to different datasets. For the second part, to judge the diversity of generated edges on dynamic graphs, we use the coefficient of variation that are different from  {Fence-GAN} \cite{fencegan}. $\beta$ shows how concerned we are about the diversity generated samples. Setting b to 10 means that we put more emphasis on the diversity of data in the loss function, which will prompt the generator to produce more diverse samples to help the discriminator to train. $CL^k$ refers to the coefficient of variation on the $K$-th dimension. Finally we uses the reciprocal of the average value of the three dimensions of $CL^k$ to measure the diversity of the data. And $CL^k$ can be formulated as follows:
\begin{equation}
CL^k = \frac{\frac{1}{N}\sum_{i=1}^N(||G(z_i^k)-\mu^k||_2)}{\mu^k}
\end{equation}
where $CL^K$ is the $k-dim$ of the interaction $(u,v,t)$, measuring the diversity of data in $k$-th dimension. And $k$ is in $[1,3]$ Through the design of this module, we can successfully generate high-quality and diverse negative samples, which lays a solid foundation for the superior performance of our model.


\subsection{Discriminator}
The purpose of the discriminator is to distinguish normal and abnormal edges from the input edges. This requires that it can capture complex continuous dynamic graph structures. Nowadays, the most powerful framework is the encoder-decoder structure. And our model design follows this architecture. where the two modules are defined as follows.
\subsubsection{Encoder}
The performance of anomaly detection models usually depends heavily on the encoding ability of the discriminator. In existing continuous dynamic graph models,  {TGNS} \cite{TGNNS} is a popular model. Based on this model,  {PINT} \cite{pint} analyzes the inherent limitations of existing messaging models and proposes to use injective temporal aggregation and positional features to overcome the limitations of temporal Weisfeiler-Leman test \cite{maron2019provably,morris2019weisfeiler}. Although existing dynamic graph models are mainly designed for link prediction tasks, models that capture dynamic graph structures are still applicable to anomaly detection tasks because they can successfully capture complex dynamic graph structures. In our task, how to build a more excellent continuous dynamic graph model is not the focus of this paper. Based on above considerations, we use the encoder structure as  {PINT} to capture complex continuous dynamic graphs.



Specifically, before the model starts training, we first calculate how many temporal paths there are between two nodes, and take this as a relative position feature into consideration in the message passing process. The detailed computing process can be found in  {PINT} \cite{pint}

In the encoding process, we use the following process to get the edge embedding of the target edge. 
\begin{equation}
E(e_{uv}(t))= concat(emb(u,t),emb(v,t))
\end{equation}
\begin{equation}
emb(u,t)=\sum_{j\in \eta_u^k([0,t])}h(s_j(t),e_{uj},r_{j\to u}^{(t)})
\end{equation}
where $h$ is a learnable function that can be designed with different forms. $s_v(t)$ are memory of $v$. $e_{uj}$, $v_u(t)$, and $v_v(t)$ is the attribute of edge and nodes, which can be set to zero if missing. By recursively aggregating the information of k-layers of neighbors, we can finally get the edge embedding.

To keep updating memory, we use the following process:
For each interaction $(u,v,t)$, we first find temporal neighbors for head and tail and then get messages from them. This process can be defined by the following formula: 
\begin{equation}
s_u(t)=upd(s_u(t^-),s_v(t^-),\Delta t,e_{uv}(t))
\end{equation}
where $upd$ is a learnable function that has different implementing ways. $s_u(t^-)$ is the memory of node $u$ just before time $t$.
Although function $h$ and $upd$ can be implemented in different forms, to ensure the performance of our detecting model, we follow the settings as  {PINT} in practice.



Although link prediction and anomaly detection have limited generality in modeling continuous dynamic graphs, the different training objectives between link prediction and anomaly detection still lead to inability to do exactly the same in model settings. In fact, in order to complete the task of link prediction, existing dynamic graph models usually follow the framework of unsupervised training, which helps model training by constructing negative samples in the training phase, and performs model evaluation through negative sampling in the testing phase. However, different training objectives and evaluation methods usually require different construction methods of negative samples with different characteristics during model training for its application in anomaly detection. 
Directly applying training methods designed for link prediction in anomaly detection may lead to suboptimal results as shown in \ref{main results}.

\subsubsection{Decoder}
The purpose of the decoder part is to generate anomaly score from the edges embedding and make judgments. Although this part can be designed to be very complex, for simplicity, we define the model as follows:
\begin{equation}
F(e)=Linear(ReLU(Linear(z(e))))
\end{equation}
Where $e$ is the input edge embedding from encoder. In our model, we use the following function as the overall error of the discriminator.
\begin{equation}
\begin{aligned}
L_D(G(z_i),x_i,D) =\frac{1}{N} \sum_{i=1}^{N} [ -\gamma \rm log(D(G(z_i)))\\
 -log(1-D(x_i)) ]
\end{aligned}
\end{equation}
Where $\gamma > 0$ is a hyperparameter determining we put how much attention on abnormal edges. When $\gamma$ is lower than 1, we put more attention on normal edges and less emphasis on anomalous edges. When $\gamma$ is greater than 1, we put more attention on anomalous edges and less emphasis on normal edges. By adjusting this hyperparameter, \ourmethod{} can adapt to various datasets.

\subsection{Training process}
\begin{algorithm}[t]
	\renewcommand{\algorithmicrequire}{\textbf{Input:}}
	\renewcommand{\algorithmicensure}{\textbf{Output:}}
	\caption{Training process of GADY}
	\label{alg1}
	\begin{algorithmic}[1]
            \STATE Preprocess: calculate the positional features $r_{\left(i\to j\right)}^{(t)}$ of each node i relative to another node j.
		\REPEAT 
            \STATE Input noise $z_i$ to generator $G$ and get fake interactions 
            \STATE Input true and fake interactions to Discriminator $D$
            \STATE For each interaction $(u,v,t)$
		\STATE For each layer $l$ :
            \STATE \quad \quad Aggregate message from temporal neighbors
		\STATE Get embedding of two nodes $u$ and $v$
		\STATE Concatenate to get edge embedding $E(e_uv(t))$ 
		\STATE Decode to get judgement $D(e_uv(t))$
		\STATE Update memory
		\STATE Pass back $L_G$ and $L_D$
		\UNTIL End
	\end{algorithmic}  
\end{algorithm}

In the training process, firstly, the generator receives noise and outputs the generated pseudo edges, then the pseudo edges and the real edges are input into the encoder for encoding to obtain the edges embedding, and finally the edges embedding is decoded into the labels to determine as anomaly or not. Finally, the loss of the generator $L_G$ and the loss of the discriminator $L_D$ passed back to update parameters. This process can be found in Algorithm 1.

\section{Experiments}
\begin{table*}[ht]
\caption{Comparison results of AUC and AP metric between different methods injecting different abnormal ratios on different datasets, where the best performance is shown in \textbf{bold} and second best performance is marked with \underline{underline}. Model
1\%, 5\%, 10\% refer to anomaly ratios respectively.}
\label{main results}
\begin{tabular}{l|c c c c c c c c c }
\toprule[1.5pt]
 \multirow{2}[3]{*}{\textbf{Method}} & \multicolumn{3}{c}{UCI} & \multicolumn{3}{c}{Bitcoin-OTC} & \multicolumn{3}{c}{Email-DNC}\\
\cmidrule(lr){2-4}\cmidrule(lr){5-7}\cmidrule(lr){8-10}
& 1\% & 5\% & 10\% & 1\% & 5\% & 10\% & 1\% & 5\% & 10\% \\
\midrule[0.75pt]
\textsc{node2vec} &0.7371 &0.7433 &0.6960 &0.7364 &0.7081 &0.6508 &0.7391 &0.7284 &0.7103\\
\textsc{Spectral Clustering} &0.6324 &0.6104 &0.5794 &0.5949 &0.5823 &0.5591 &0.8096 &0.7857 &0.7759\\
\textsc{DeepWalk} &0.7514 &0.7391 &0.6979 &0.7080 &0.6881 &0.6396 &0.7481 &0.7303 &0.7197\\
\cmidrule(lr){1-10}
\textsc{NetWalk} &0.7758 &0.7647 &0.7226 &0.7785 &0.7694 &0.7534 &0.8105 &0.8371 &0.8305\\
\cmidrule(lr){1-10}
\textsc{TGN} &0.8771 &8667 &0.8539 &0.9411 &0.9284 &0.9196 &0.9677 &0.9474 &0.9326\\
\textsc{PINT} &0.9265 &0.9232 &0.9229 &0.9227 &0.9226 &0.9225 &0.9758 &0.9754 &0.9783\\
\cmidrule(lr){1-10}
\textsc{AddGraph} &0.8083 &0.8090 &0.7688 &0.8341 &0.8470 &0.8369 &0.8393 &0.8627 &0.8773\\
\textsc{StrGNN} &0.8179 &0.8252 &0.7959 &0.9012 &0.8775 &0.8836 &0.8775 &0.9103 &0.908\\
\textsc{TADDY} &0.8912 &0.8398 &0.837 &0.9455 &0.934 &0.9425 &0.9348 &0.9257 &0.921\\
\cmidrule(lr){1-10}
\textsc{GADY} (No-GAN) &\textbf{0.9658} &\textbf{0.9655} &\textbf{0.9656} &\textbf{0.9826} &\underline{0.9836} &\underline{0.9828} &\underline{0.9636} &\underline{0.9696} &\underline{0.9648}\\
\textsc{GADY} (GAN) &\underline{0.9600} &\underline{0.9585} &\underline{0.9597} &\underline{0.9819} &\textbf{0.9854} &\textbf{0.9839} &\textbf{0.9796} &\textbf{0.9835} &\textbf{0.9827}\\
\midrule[1pt]
\textsc{TADDY}(AP) &0.1760 &0.3428 &0.4743 &0.1402 &0.4287 &0.5995 &0.1046 &0.2854 &0.3986\\
\textsc{GADY} (No-GAN)(AP) &\underline{0.4811} &\underline{0.4801} &\underline{0.4784} &\underline{0.6105} &\underline{0.6093} &\underline{0.6221} &\underline{0.6149} &\underline{0.6617} &\underline{0.6173}\\
\textsc{GADY} (GAN)(AP) &\textbf{0.5025} &\textbf{0.6771} &\textbf{0.7617} &\textbf{0.6842} &\textbf{0.8395} &\textbf{0.8797} &\textbf{0.7462} &\textbf{0.8583} &\textbf{0.8861}\\

\bottomrule[1.5pt]
\end{tabular}
\end{table*}

\subsection{Experiment Settings}
\subsubsection{Evaluation Protocol} \textbf{AUC (Area Under Curve)} is a metric for evaluating the performance of binary classification models. It takes account of the evaluation ability of both positive and negative classes, and is widely used in anomaly detection models.

\textbf{AP (Average Precision)} is a metric for evaluating the performance of object detection models on a single class. It is based on the area under the Precision-Recall (PR) curve, which shows the precision and recall of the model at different thresholds. 

Different from AUC, AP comprehensively considers the accuracy and recall rate, which reflect whether the model can find all the anomalies and whether the anomalies found are correct. Therefore, we believe that the AP metric can comprehensively measure the detection ability of the anomaly detection model better, which is seldom considered in the previous works \cite{netwalk,addgraph,strgnn,taddy}. In our follow-up experiments, we also proved the superior performance of our model on this evaluating metric.

\subsubsection{Datasets}
To test the performance of our model, we evaluated our framework on three datasets: \textbf{UCI Message}\footnote{\url{http://konect.cc/networks/opsahl-ucsocial}} \cite{UCI}, \textbf{Email-DNC}\footnote{\url{http://networkrepository.com/email-dnc}} \cite{email-dnc}, and  \textbf{Bitcoin-OTC}\footnote{\url{http://snap.stanford.edu/data/soc-sign-bitcoin-otc}} \cite{Bitcoin-OTC}. These datasets are constructed from data from forum posts, emails, and trading systems respectively, details of which can be found in the appendix.


\subsubsection{Baselines}
We selected four classes of models as our baselines: traditional non-deep learning methods  {Node2vec} \cite{node2vec}  {Spectral Clustering} \cite{spetralclustering}  {DeepWalk} \cite{deepwalk}, methods using a combination of deep and non-deep learning  {NetWalk},\cite{netwalk} continuous dynamic graph modeling methods directly used for anomaly detection  {TGNs} \cite{TGNNS}  {PINT} \cite{pint}, and anomaly detection methods using deep learning  {AddGraph} \cite{addgraph}  {StrGNN} \cite{strgnn}  {TADDY} \cite{taddy}. A detailed introduction to baselines can be found in the appendix.

\subsubsection{Experiment Setup}
In order to evaluate the performance of our model for anomaly detection and make a fair comparison with baselines, we follow the method test set is built in  {TADDY} \cite{taddy}. Specifically, we first perform spectral clustering on the whole graph, then randomly select nodes belonging to different categories, and remove node pairs that are duplicated with the original dataset. Then, we randomly generate timestamps for node pairs within the time range of the test set, and finally add the generated pseudo-edges to the test set and sort them by time. 

In our experiments, the message function use the identity function used in  {TGN} \cite{TGNNS} Number of layers is set to 2 and the training ratio is set to 0.5. Other detailed implement details can be found in appendix.

Finally, we implement our method using PyTorch 1.12.1 \cite{paszke2019pytorch}. All experiments are performed on a Linux server with a 2.30GHz Intel(R) Xeon(R) Silver 4316 CPU, and NVIDIA Tesla-V100S GPU with 32GB memory.

\begin{figure}[t]
\begin{center}
    \centering
    \includegraphics[width=1\linewidth]{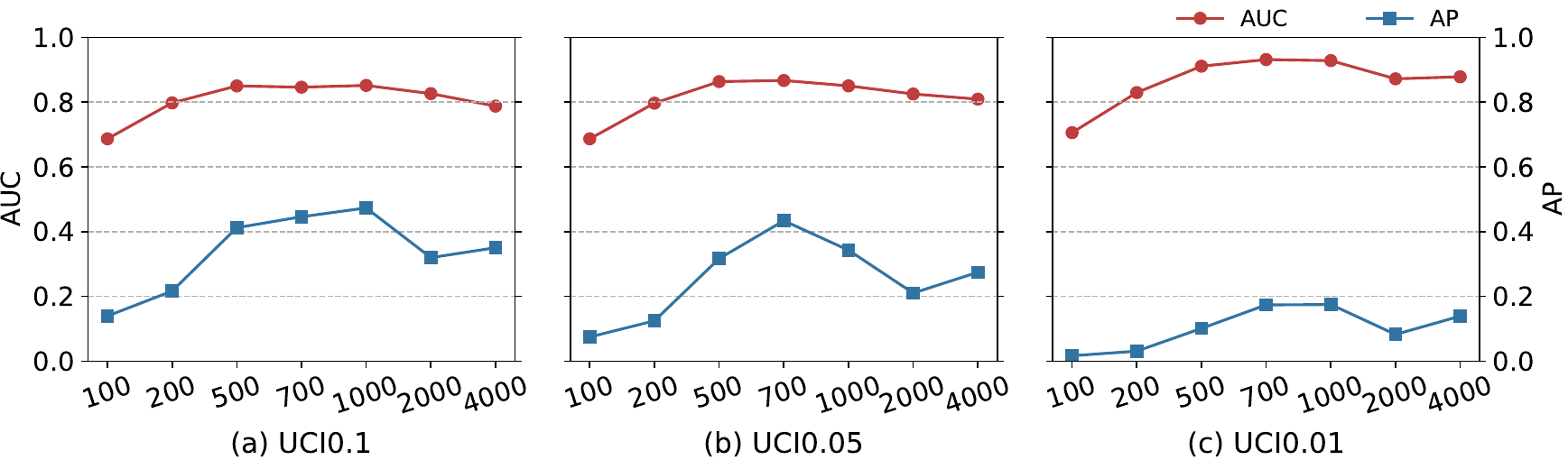}
    \end{center}
    \caption{Effects of Slice Size on detection performance. Too small or too large slice size will limit model performance. So it is difficult for discrete dynamic methods to capture fine-grained time information, which is an fatal flaw of this kind of method and validate the superiority of continuous dynamic methods.}
\label{fig:size study}
\end{figure}

\begin{figure}[t]
    \includegraphics[width=1\linewidth]{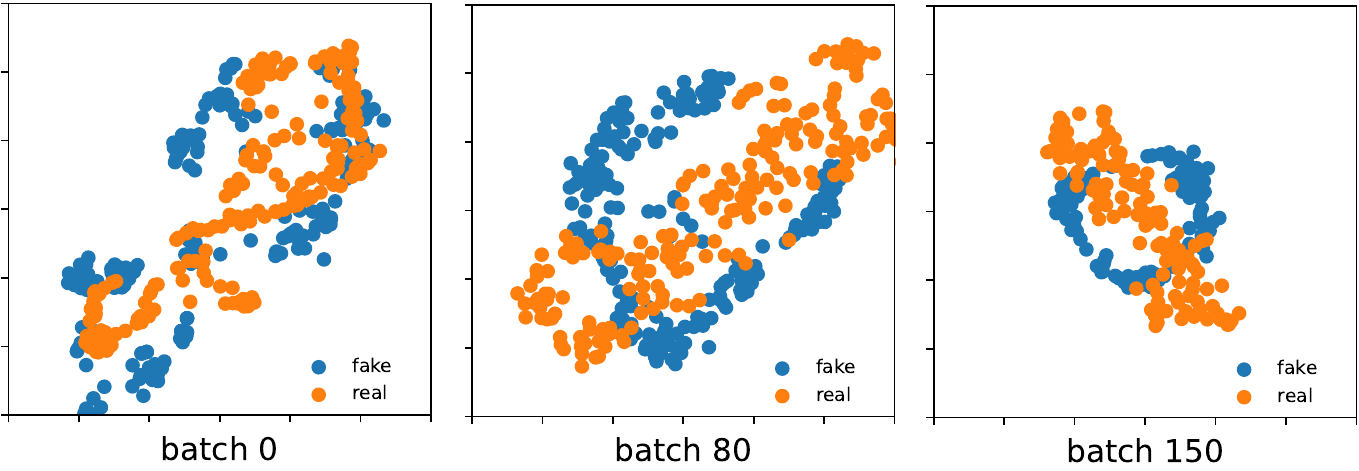}
    \caption{Comparison of negative sample distributions generated in different periods in the second epoch.It can be seen that with the training of the model, the generated negative samples gradually gather evenly near the real samples, which proves that our generator can indeed generate diverse and high-quality negative samples.}
\label{distribution}
\end{figure}

\subsection{Main Results}
We tested our model on three datasets and set the injected anomaly ratio to 1\%, 5\%, and 10\% respectively. We also show the results without using the GAN model (by using the same training method as TADDY to train the discriminator), and the results of the complete model using GAN.
The experimental results are shown in Table \ref{main results}. From the experimental results, we summarize the following conclusions: 

\begin{itemize}
\item Our model has excellent results, and the use of the GAN module can greatly improve the ability of the model to detect anomalies. Our model outperforms state-of-the-art method by 14.6\% at most on the UCI dataset with anomaly ratio of 10\%, achieving SOTA on three real world datasets, and the use of the GAN module can greatly improve the ability of the model to detect anomalies which is shown in AP metric.

\item Continuous dynamic graph models have remarkable advantages over discrete dynamic graphs. It can be seen from the effect of the GADY model that does not use GAN that it significantly surpasses the ordinary discrete dynamic graph, which verifies our idea.

\item Our model performance is insensitive to anomaly ratios. From the table, it can be found that other models are more sensitive to abnormal injection ratio. However, our method remains a high performance regardless of what the anomaly ratio is.This will allow our model to have a wider range of applicability to different anomaly ratios.

\item The idea of directly using continuous dynamic graph models for anomaly detection tasks is not enough. From the table, we can find that the detection effects of  {TGN} \cite{TGNNS} and  {PINT} \cite{pint} are similar to the traditional anomaly detection model TADDY, but still have a large distance compared with our model. This also verifies our conjecture: although continuous dynamic graph models for different tasks are all designed for capturing the dynamic graph structure, the model for link prediction tasks cannot be directly applied to anomaly detection. It needs to use different negative sampling methods in order to achieve better results.
\end{itemize}

\subsection{Slice Size Study}
In order to further validate the superiority of continuous dynamic graph anomaly detection compared to descrete graph processing, we adjust the slice size of to 100, 300, 500, 700, 1000, 1500, 2000, and 5000 respectively, and test  {TADDY} \cite{taddy} on the UCI dataset with anomaly rate 1\%, 5\%, 10\%. The experimental results are shown in Figure \ref{fig:size study}.

Some people may say that a discrete form of dynamic graph can capture more fine-grained temporal information by increasing the number of slices, but this is unrealistically verified in this experiment.


From the experimental results, no matter whether the slice size is large or small, the model effect cannot reach the best. When the slice size is too small, it will be difficult for GNNs to capture the structural information on the slice; when the slice size is too large, too much time information will be ignored, thus limiting the anomaly detection effect of discrete dynamic graphs.

\begin{figure}[t]
    \includegraphics[width=1\linewidth]{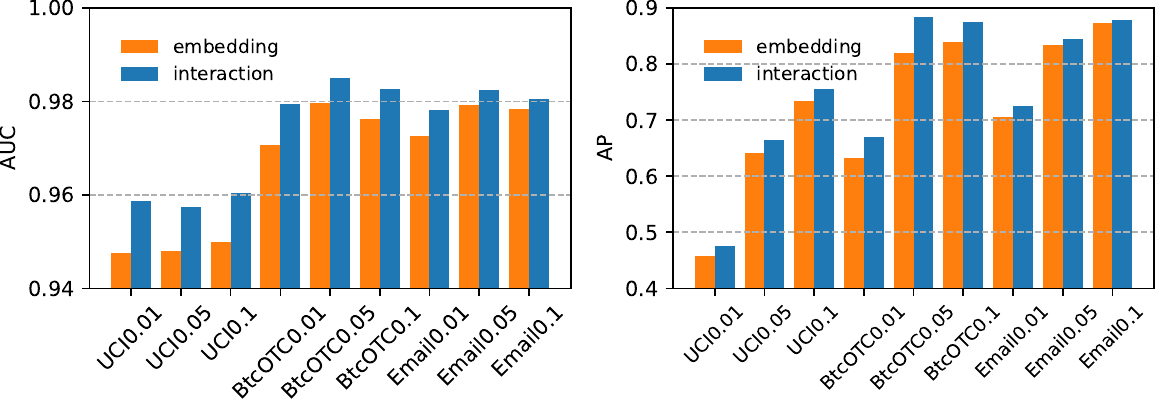}
    \caption{Comparison between generating interactions or embeddings. The results shows that generating interactions has better performance than generating embeddings directly.}
\label{fig:interaction1}
\end{figure}


\subsection{Performance of Generated Negative Edges}
This experiment is to examine the performance of the generator on anomalous samples. Specifically, we select the anomalous samples generated in different batches in the second epoch, and visualized the data distribution results as shown in the Figure \ref{distribution}. Form the results, we find that with the continuous training of the model, the generated samples are gathered at the boundary of the real samples, which proves the high quality of the samples generated by the generator, and the generated samples are evenly distributed around the real samples, which demonstrates the diversity of samples generated by the generator. 

\subsection{Generate edges or edge Embedding}
In this experiment, we test the model performance of using the generator to directly generate edge embedding or edge interactions, which is shown in Figure \ref{fig:interaction1}. From the results, we can find that no matter which dataset and what the abnormality ratio is, the results of directly generating interactions are always better than those directly generating edge embedding. This may be because the method of directly generating edge encoding ignores the encoder's process of modeling the interaction, thus causing more bias. Therefore, we can conclude that using a generator to generate interactions is a better choice than directly generating edge representations.






\section{Conclusion}
In this paper, we discovered the shortcomings of existing discrete dynamic graphs for anomaly detection, and proposed that continuous dynamic graphs should be used for anomaly detection tasks. Also, our study discovers how existing continuous dynamic graphs for link prediction should perform on anomaly detection tasks. On this basis, we explored how to use the GAN model to obtain more high-quality and diverse samples. The final experimental results not only prove the excellent performance of the continuous dynamic graph in anomaly detection, but also prove the role of the GAN module in helping the model identify anomalies. Supplementary experiments on Slice Size, visualization of generated samples, generated edges or edge embedding further prove the shortcomings of discrete dynamic methods, the superiority of GAN to generate negative samples, and the superiority of model settings.



\bibliography{aaai24}

\begin{appendices}

\section{Appendix}

\subsection{Complexity Analysis}
\begin{figure}[h]
\begin{centering}
    \centering
    \includegraphics[width=1\linewidth]{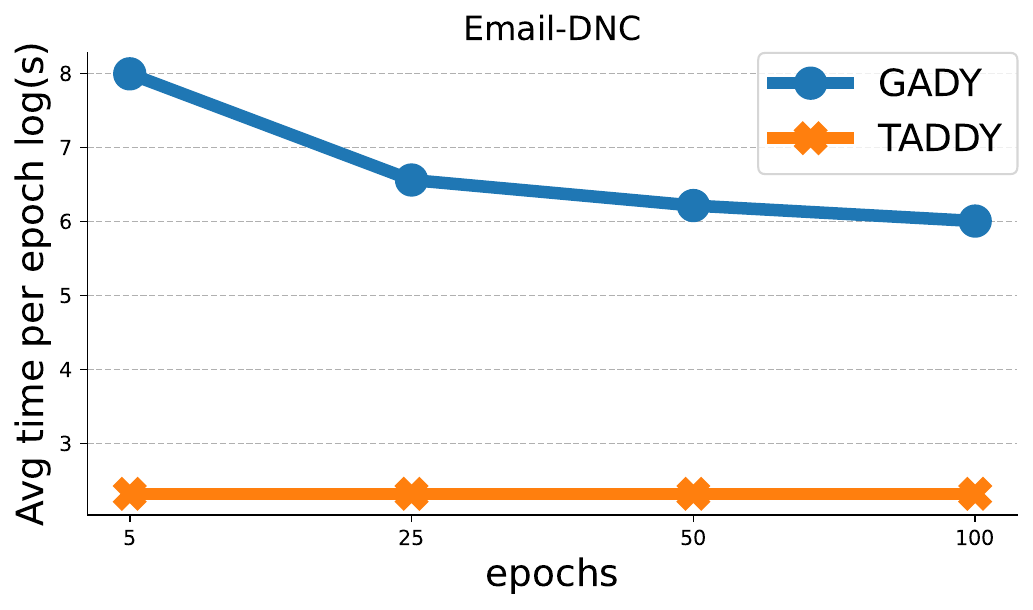}
    \end{centering}
    \caption{.}
\label{fig:time compare}
\end{figure}

In the comparison of time complexity, we conducted a comparative experiment, and the result are shown in Figure \ref{fig:time compare}. The result shows that overall our GADY is slower than TADDY, and the running time of each epoch is getting shorter due to the preprocessing time, but the overall is still within an acceptable range. It is worth mentioning that if the model of the encoder part can be replaced with a lighter structure, the time overhead will be much lower. And the specific time complexity will be extremely dependent on the model selected by the encoder part.

\subsection{Datasets}
\begin{itemize}

\item \textbf{UCI Message}\footnote{\url{http://konect.cc/networks/opsahl-ucsocial}} \cite{UCI} is a social network collected from a forum of the University of California, Irvine. It has 1,899 nodes and 13,838 edges where each node represents a student and an edge represents a message sent.

\item \textbf{Email-DNC}\footnote{\url{http://networkrepository.com/email-dnc}} \cite{email-dnc} is an email network from the 2016 Democratic National Committee email leak. It has 1,866 nodes and 39264 edges where each node represents a person in the Democratic Party and an edge represents an email sent from one person to another.

\item \textbf{Bitcoin-OTC}\footnote{\url{http://snap.stanford.edu/data/soc-sign-bitcoin-otc}} \cite{Bitcoin-OTC} is a network of who trusts whom among users who trade on the Bitcoin platform. It has 5,881 nodes and 35,588 edges where nodes are users and edges are ratings between users.
\end{itemize}

\subsection{Baselines}
We compared GADY with nine competitive methods, which mainly include two categories: deep learning methods and graph encoding methods.
\begin{itemize}

\item \textbf{DeepWalk} \cite{deepwalk} is a social network collected from a forum of the University of California, Irvine, where each node represents a student and an edge represents a message sent.

\item \textbf{Node2vec} \cite{node2vec} is a social network collected from a forum of the University of California, Irvine, where each node represents a student and an edge represents a message sent.

\item \textbf{Spectral Clustering} \cite{spetralclustering} is a social network collected from a forum of the University of California, Irvine, where each node represents a student and an edge represents a message sent.

\item \textbf{Netwalk} \cite{netwalk} is a classic anomaly detection method on dynamic graphs, which generates node encodings based on random walk and dynamically updates reservoirs to model networks, and finally uses a dynamic clustering model to score edges for anomalies.

\item \textbf{TGN} \cite{TGNNS} is a classic framework for deep learning on dynamic graphs represented as sequences of timed events. It uses memory modules and graph-based operators to model patterns of dynamic systems and is able to significantly outperform previous approaches being at the same time more computationally efficient. 

\item \textbf{PINT} \cite{pint} is a novel architecture that leverages injective temporal message passing and relative positional features to boost the expressive power of TGN. We use this baseline to demonstrate the strong expressive power of our model.

\item \textbf{AddGraph} \cite{addgraph} is an end-to-end dynamic graph anomaly detection model. It uses GCN and GRU-attention to capture structural and temporal information respectively.

\item \textbf{StrGNN} \cite{strgnn} is an end-to-end model for anomaly detection on dynamic graphs. It first takes h-hop subgraphs for the target edge, then uses GCN model to model the structural information within the slice, and uses GRU to model the temporal information between different slices.

\item \textbf{Taddy} \cite{taddy} is an end-to-end model for anomaly detection on dynamic graphs. It first samples subgraphs and obtains global encodings by diffusion, obtains local encodings based on distance on subgraphs and obtains relative time encodings according to time difference. Then it uses transformers to model dynamic graph networks and then uses a scoring module to obtain anomaly scores.
\end{itemize}

\subsection{Experiment Setup}
We set alpha to 0.1, beta to 15, and gamma to 0.1 for experiments. For the parameters setting of the encoder part, we followed the hyperparameters setting discussed in PINT\cite{pint}. For all hyperparameters settings in the baselines, we also follow the best parameters proposed in the papers for setting.

\end{appendices}
\end{document}